\documentclass[10pt,twocolumn,letterpaper]{article}

\usepackage{wacv}
\usepackage{times}
\usepackage{epsfig}
\usepackage{graphicx}
\usepackage{amsmath}
\usepackage{amssymb}
\usepackage{authblk}

\usepackage{caption}
\usepackage{float}
\usepackage[noend]{algpseudocode}

\usepackage{color}
\usepackage[linesnumbered,ruled,vlined]{algorithm2e}
\SetKwInput{KwInput}{Input}                % Set the Input
\SetKwInput{KwOutput}{Output}              % set the Output
\usepackage{subcaption}
\usepackage{booktabs}
\usepackage{multirow}
\usepackage{array}

\def\wacvPaperID{896} % Enter the WACV Paper ID here

\wacvfinalcopy % *** Uncomment this line for the final submission
\ifwacvfinal
\def\assignedStartPage{1} % *** Enter the assigned starting page number (instead of 9876)
\fi

\ifwacvfinal
\usepackage[breaklinks=true,bookmarks=false]{hyperref}
\else
\usepackage[pagebackref=true,breaklinks=true,colorlinks,bookmarks=false]{hyperref}
\fi

\ifwacvfinal
\else
\fi

\begin{document}

%%%%%%%%% TITLE
\title{EvidentialMix: Learning with Combined Open-set and Closed-set Noisy Labels}

% \author{Ragav Sachdeva\\
% University of Adelaide\\
% % For a paper whose authors are all at the same institution,
% % omit the following lines up until the closing ``}''.
% % Additional authors and addresses can be added with ``\and'',
% % just like the second author.
% % To save space, use either the email address or home page, not both
% \and
% Filipe R. Cordeiro\\
% Universidade Federal Rural de Pernambuco\\

% % \and
% % Filipe R. Cordeiro\\
% % Universidade Federal Rural de Pernambuco\\

% \and
% Vasileios Belagiannis\\
% Universität Ulm\\

% \and
% Ian Reid\\
% University of Adelaide\\

% \and
% Gustavo Carneiro\\
% University of Adelaide\\

% }
\author[1]{Ragav Sachdeva}
\author[2]{Filipe R. Cordeiro}
\author[3]{Vasileios Belagiannis}
\author[1]{Ian Reid}
\author[1]{Gustavo Carneiro}
\affil[1]{University of Adelaide, Adelaide, Australia}
\affil[2]{Universidade Federal Rural de Pernambuco, Recife, Brazil}
\affil[3]{Universität Ulm, Ulm, Germany}
% \author[2]{Filipe R. Cordeiro}
% \address{University of Adelaide}
% \IEEEauthorblockA{University of Adelaide}

\maketitle

\begin{abstract}
%Growing applications of Deep Neural Networks (DNNs) have made them a captivating tool for researchers and academics alike. However, these DNNs typically require large-scale data sets with clean and reliable data to train them, which are not always available for a new domain. Acquiring such large-scale data sets with precise annotations is very expensive and time-consuming, and the cheap alternatives often yield data sets that have noisy labels.
% Considering the recent trend of employing these deep learning models in the real world and the possible implications it may have on the society, this work aims to research ways to make DNNs more robust to such noisy data sets.

The efficacy of deep learning depends on large-scale data sets that have been carefully curated with reliable data acquisition and annotation processes.
However, acquiring such large-scale data sets with precise annotations is very expensive and time-consuming, and the cheap alternatives often yield data sets that have noisy labels.
The field has addressed this problem by focusing on training models under two types of label noise: 1) closed-set noise, where some training samples are incorrectly annotated to a training label other than their known true class; and 2) open-set noise, where the training set includes samples that possess a true class that is (strictly) not contained in the set of known training labels.
% Both types of label noise mentioned above have always been addressed separately in the field, which is arguably limiting when emulating real-life noisy data sets.
%The closed-set noise problem is typically addressed by identifying and re-labelling the noisy label samples, so the model can be re-trained with the updated training set. On the other hand, open-set noise is handled by identifying and reducing the weight of these samples in the training process.
In this  work, we study a new variant of the noisy label problem that combines the open-set and closed-set noisy labels, and introduce a benchmark evaluation to assess the performance of training algorithms under this setup.
We argue that such problem is more general and better reflects the noisy label scenarios in practice.
Furthermore, we propose a novel algorithm, called EvidentialMix, that addresses this problem and compare its performance with the state-of-the-art methods for both closed-set and open-set noise on the proposed benchmark. Our results show that our method produces superior classification results and better feature representations than previous state-of-the-art methods. The code is available at {\color{red}\url{ https://github.com/ragavsachdeva/EvidentialMix}}.
% Noisy label training methods assume a fixed set of training labels and training samples that can come from one or more training domains.
% While closed-set noisy label learning mixes correct and incorrect labels for training data samples coming from a single domain, open-set noisy label approaches mix data and correct labels from one domain with samples and incorrect labels from another domain.
% The closed-set noisy label problem has been traditionally addressed by identifying the noisy label samples that are then re-labelled or un-labelled, so the model can be re-trained with the updated training set.
% On the other hand, open-set approaches aim to identify and remove the noisy label samples from the training process given the training labels may not represent well those samples.
% In this paper, we propose the study of a new noisy label problem that combines open-set and closed-set noisy labels.  Such problem is more general and potentially more useful when dealing with real-life noisy label scenarios.
% Furthermore, we propose an algorithm that addresses this new problem by extending a state-of-the-art (SOTA) noisy label learning method using a loss based on the theory of subjective logic to handle both the closed-set and open-set noise problems.
% We introduce benchmark experiments for this new problem and compare the performance of closed-set and open-set SOTA approaches with our proposed approach.  
% Results show that our method produces superior results than SOTA noisy label learning methods. Benchmark experiments and code will be available upon paper acceptance.
\end{abstract}

%%%%%%%%% BODY TEXT
\section{Introduction}
\label{sec:intro}

% why noisy label learning?
Deep learning has achieved outstanding results in several important classification problems, using large and well-curated training data~\cite{DLReview,deng2009imagenet}.
However, most of the interesting data sets available in the society are orders of magnitude larger, but poorly curated, which means that the data may contain acquisition and labeling mistakes that can lead to poor generalisation~\cite{zhang2016understanding}.
% \gustavo{insert a figure that shows an image search that shows different domains and wrong labels.}
Therefore, one of the important challenges of the field is the development of methods that can cope with such noisy label data sets. Lately, researchers have greatly fostered the development of this field by studying controlled synthetic label noise and discovering theories or methodologies that can then be applied to real-world noisy data sets.

The types of label noise investigated thus far can be classified into two categories -- \textit{closed-set} and \textit{open-set} noise. Although these terms (`closed-set' and `open-set') were coined only recently by Wang et al. in \cite{OpenSetPaper} where they introduced the open-set noisy label problem, the closed-set noisy label problem has been extensively studied since much before.
% types of noisy label -- open and closed set, but no combined problem -- we REALLY NEED TO SHOW THE IMPORTANCE OF THE PROBLEM HERE
When handling closed-set label noise, the majority of the learning algorithms assume a fixed set of training labels~\cite{DivideMix,SELF}. In this setting, some of the training samples are annotated to an incorrect label, while their true class is present in the training label set.  
These mistakes can be completely random, where labels are flipped arbitrarily to an incorrect class, or consistent, when the annotator is genuinely confused about the annotation of a particular sample. A less studied label noise is the open-set noisy label problem~\cite{OpenSetPaper}, where we incorrectly sample some data observations, such that their true annotation is \textit{not} contained within the set of known training labels. A hyperbolic example of such a setting could be the presence of a \textit{horse} image in the training set for modelling a \textit{cats vs dogs} binary classifier. As evident from their definitions, these two types of label noise are mutually exclusive, i.e., a given noisy label cannot be closed-set and open-set at the same time.

\begin{figure}
\centering
  \includegraphics[width=1.0\linewidth]{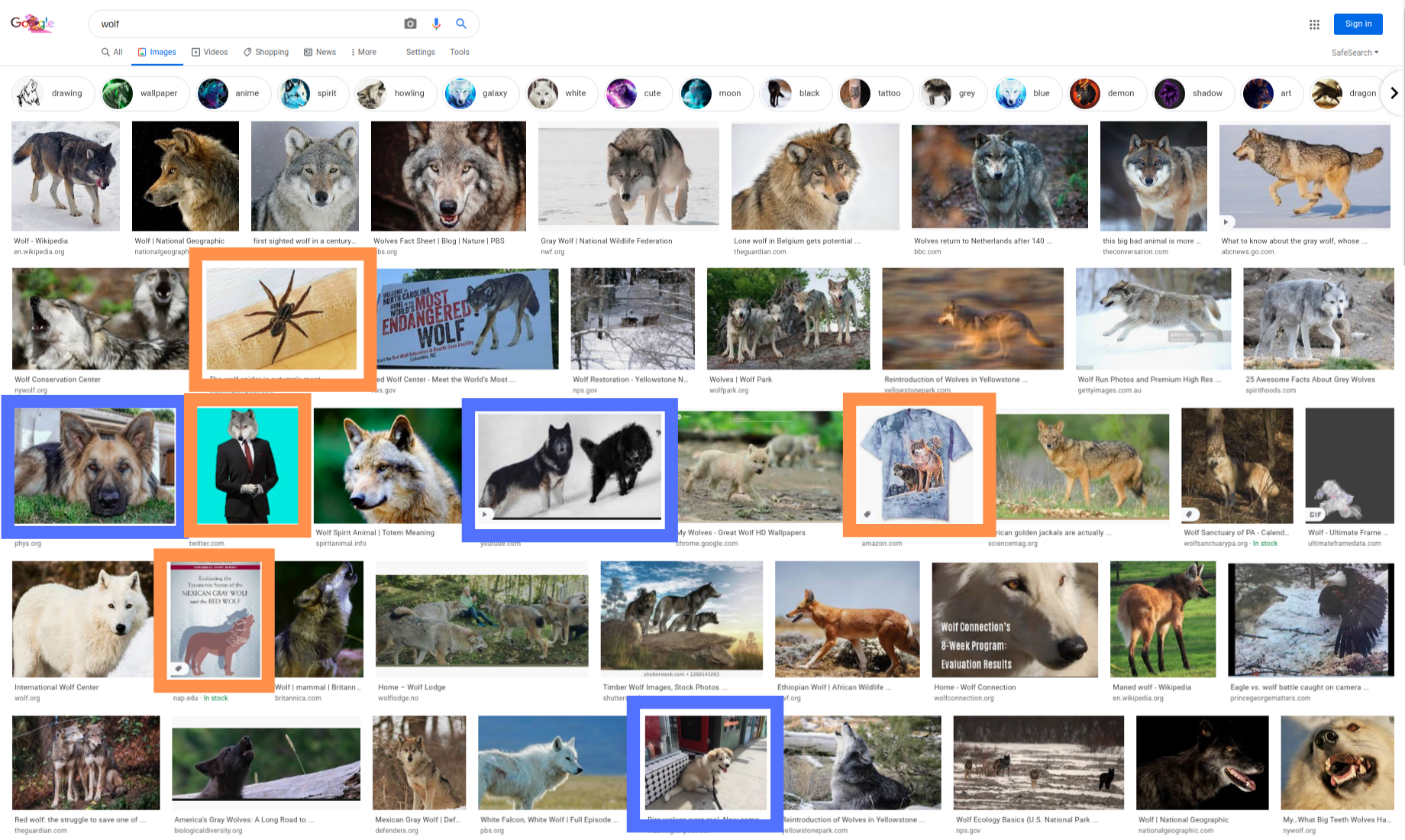}
  \caption{Results of a search engine query to collect data for a wolf-vs-dog binary classifier. The search keyword used here is ``wolf". The images bounded by an orange box are open-set noise (i.e. neither wolf nor dog) and the ones bounded by a blue box are closed-set noise (i.e. labelled as wolf but are actually a dog).}
  \label{fig:wolfvsdog}
\end{figure}

It is quite easy to substantiate that both open-set and closed-set noise are likely to co-occur in real-world data sets.
For instance, recent methods for large scale data collection propose the use of querying commercial search engines (e.g., Google Images), where the search keywords serve as the labels of the queried images. It is evident from Figure~\ref{fig:wolfvsdog} that collecting images using such methods can lead to both open-set and closed-set noise.
However, thus far, no systematic study with \textit{controlled} label noise has been presented, where the training data set contains both types of label noise simultaneously. Even though there have been papers that evaluated their proposed methods on both~\cite{lee2019robust, OpenSetPaper}, the training data sets have been exclusively corrupted with either closed-set noise or open-set noise, but never in a combined fashion.

% proposal, main results and conclusion punch line
In this paper, we formulate a novel benchmark evaluation to address the noisy label learning problem that consists of a combination of closed-set and open-set noise. This proposed benchmark evaluation is defined by three variables: 1) the total proportion of label noise in the training set, represented by $\rho \in [0,1]$; 2) the proportion of closed-set noise \textit{within the set of samples containing noisy labels}, denoted by $\omega \in [0,1]$ (this implies that $\rho \times \omega \%$ samples of the entire data set have a closed-set noisy label and $\rho \times (1 - \omega) \%$ samples of the entire data set have an open-set noisy label); and 3) the source of open-set noisy label data.
% (in particular, we use CIFAR-10~\cite{krizhevsky2009learning} 
% as the main training set, and CIFAR-100~\cite{krizhevsky2009learning} and ImageNet32~\cite{chrabaszcz2017downsampled} for the open-set noise).
% \ragav{I don't like the idea of the choice of open-set dataset being a "third" variable. This somehow feels restricting our method to only CIFAR100 and Imagenet32 when that is not the case.}
Note that this setup generalises both types of label noise as it can collapse to one of the two noise types when $\omega \in \{0,1\}$.

\begin{figure}
  \includegraphics[width=\linewidth]{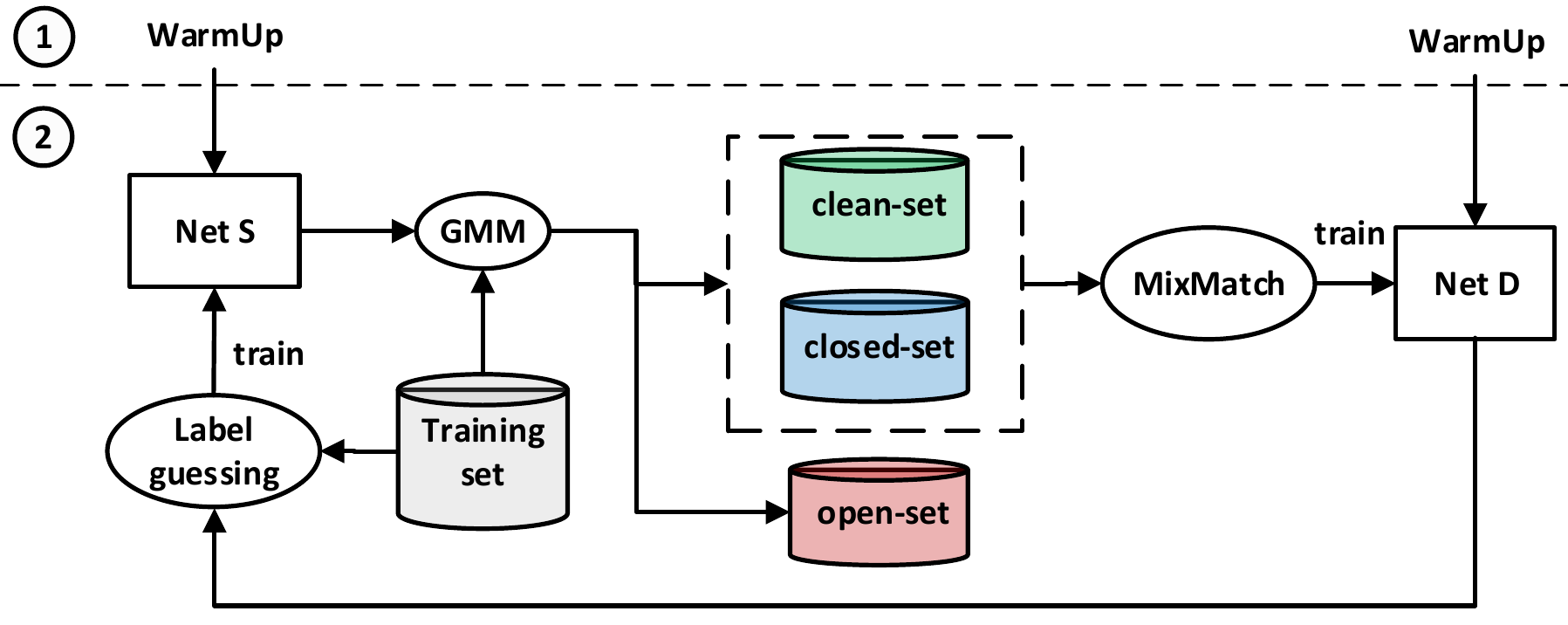}
  \caption{Our proposed method, EvidentialMix, relies on two models, NetD and NetS.  Both models are initially trained for a few epochs (see (1) WarmUp) using just a simple training mechanism that does not include any approach to handle label noise. Next in (2), we fit a $\psi$-component Gaussian Mixture Model (GMM) to the loss distribution of the samples from NetS in order to separate the training set into clean, closed-set and open-set samples. Following this, NetD is trained with the SSL learning mechanism~\cite{DivideMix} using the predicted clean and closed-set samples only. Finally, NetS is trained on the entire training set to minimise the subjective logic loss~\cite{sl} using the labels estimated from NetD, and the process in (2) repeats.}
  \label{fig:diagram}
\end{figure}

The state-of-the-art (SOTA) approaches that aim to solve the closed-set noisy label problem focus on methods that identify the samples that were incorrectly annotated and update their labels with semi-supervised learning (SSL) approaches~\cite{DivideMix} for the next training iteration.  
% % When samples are re-labelled, the training set samples are updated with the new labels~\cite{}, and when samples are un-labelled, the training is based on semi-supervised learning (SS)~\cite{}.
This strategy is likely to fail in the open-set problem because it assumes that there exists a correct class in the training labels for every training sample, which is not the case. On the other hand, the main approach addressing the open-set noise problem targets the identification of noisy samples to reduce their weights in the learning process~\cite{OpenSetPaper}.  Such strategy is inefficient in closed-set problems because the closed-set noisy label samples are still very meaningful during the  SSL stage. Hence, to be robust in the scenarios where both closed-set and open-set noise samples are present, the learning algorithm must be able to identify the type of label noise affecting each training sample, and then either update the label, if it is closed-set noise, or reduce its weight, if it is open-set noise. To achieve this, we propose a new learning algorithm, called EvidentialMix (EDM) -- see Fig.~\ref{fig:diagram}. 
The key contributions of our proposed algorithm are the following:
\begin{itemize}
  \item EDM is able to accurately distinguish between clean, open-set and closed-set samples, thus allowing it to exercise different learning mechanisms depending on the type of label noise. In comparison, previous methods~\cite{DivideMix, OpenSetPaper} can only separate clean samples from noisy ones, but not closed-noise from the open-noise samples.
  \item We show that our method can learn superior feature representations than previous methods as evident from the t-SNE plot in Figure~\ref{fig:tsne}, where our method has a unique cluster for each of the known label/class and another \textit{separate} cluster for open-set samples. In comparison, previous methods have shown to largely overfit the open-set samples and incorrectly cluster them to one of the known classes.
  \item We experimentally show that EDM produces classification accuracy that is comparable or better than the previous methods on various label noise rates (including the extreme case where $\omega \in \{0,1\}$).
\end{itemize}
% EDM combines DivideMix~\cite{DivideMix}, which is the current closed-set SOTA method, with a loss function based on the theory of subjective logic~\cite{sl}, which captures classification uncertainty to detect closed-set and open-set noisy samples -- see Fig.~\ref{fig:diagram}.
% More specifically, the loss function used by DivideMix~\cite{DivideMix} can only distinguish between samples that have low loss values (assumed to be correctly labelled) and samples with high-loss values (assumed to have a wrong label) -- this makes the classification between open-set and closed-set samples challenging because they are mixed in the same set of high-loss value samples.
% Subjective logic loss~\cite{sl}, on the other hand, enables the classification of the two types of loss above in addition to the classification of a mid-range loss value associated with open-set noisy label samples.
% Results from the proposed benchmark comparing the performance of our method with 
% the current closed-set~\cite{DivideMix,lee2019robust} and open-set~\cite{OpenSetPaper,lee2019robust} SOTA approaches show that we have the most accurate classification in our proposed benchmark evaluation. 
%\vb{we could add 1-2 more sentences to describe our approach.}
%\gustavo{we need another figure that shows the main stages of our approach}

%impressive results on problems that seemed insurmountable, if not impossible, not too long ago. Its growing applications in areas ranging from computer vision and natural language processing to bioinformatics and medical imaging has made it a very captivating tool for industry and academics alike \cite{DLReview}. However, many of these tasks require large-scale datasets with clean and reliable data to train deep neural networks (DNNs), which are not always available for a new domain. In addition, acquiring such large-scale datasets with precise annotations is very expensive and time-consuming, thus limiting the use of DNNs in real-world settings, despite their potential. 

%There are cheap but imperfect alternatives to collect data and annotations at large scale such as querying commercial search engines \cite{WebVision}, crowd-sourcing from non-experts etc., particularly for image data where the tags and online query keywords are treated as valid labels. This, as one can imagine, inevitably yields datasets that contain noisy or miSabelled samples. Furthermore, most of the “clean datasets” are from humans, who are liable to make mistakes.

%Considering the recent trend of employing these deep learning models in the real world and the possible implications it may have on the society, this work aims to research ways to make DNNs more robust to such noisy datasets.

%-------------------------------------------------------------------------
\section{Prior Work}
\label{sec:lit_review}

There is an increasing interest in the study of modelling deep learning classifiers with noisy labels.  
For the closed-set noise, Reed et al.~\cite{ReedEarly} proposed one of the first approaches that uses a transition matrix to learn how labels switch between different classes.  
%This method did not explicitly try to identify the samples containing noisy labels, which is an important component of the SOTA methods. 
The use of transition matrices has been further explored in many different ways~\cite{patrini2017making,goldberger2016training}, but none of them show competitive results, likely because they do not include mechanisms to identify and handle samples containing noisy labels.
Data augmentation approaches~\cite{zhang2017mixup} have been successively explored by closed-set noisy label methods, where the idea is that it can naturally increase the training robustness to label noise.
%and they are now part of most of SOTA approaches~\cite{DivideMix,SELF}.
Meta-learning is another technique explored in closed-set noisy label problems~\cite{li2019learning,ren2018learning}, but the need for clean validation sets or artificial new training tasks makes this technique relatively unexplored.
%performed early works on learning robustly under label noise for deep neural networks. Following that Rolnick et al. showed that for classification tasks deep neural networks come with natural robustness to label noise following a particular random distribution. No modification of the network or the training procedure is required to achieve this robustness \cite{RolnickRobust}. 
The use of curriculum learning (CL)~\cite{jiang2018mentornet} for closed-set problems has been explored to re-label training samples dynamically during training, based on their loss values. This approach has been extended with the training of multiple models~\cite{malach2017decoupling,yu2019does} that aim to focus the training on samples with small loss that are inconsistently classified by the multiple models.  
%Similar CL approaches are also present in SOTA methods~\cite{DivideMix}.
Recently, the explicit identification of noisy samples using negative learning has been explored by Kim et al.~\cite{kim2019nlnl}, with competitive results.  
Another important approach in handling label noise is model ensembling, as proposed by Tarvainen et al.\cite{tarvainen2017mean}.
The use of robust generative classifiers (RoG) to improve the performance of discriminative classifiers has been explored by Lee et al.~\cite{lee2019robust}, where they build an ensemble of robust linear discriminative models using features extracted from several layers of the trained discriminative model -- in principle, this approach has the potential to improve the performance of any method and has been successively tested in closed-set and open-set scenarios.

The learning with open-set noisy labels has only recently been explored by Wang et al.~\cite{OpenSetPaper}, where the idea  is to identify the samples containing noisy labels and reduce their weight in the training process since they almost certainly belong to a class not represented in the training set. Given that they are the only ones to explicitly address the open-set noise, their method is the main baseline for that problem.
%\gustavo{we need a better summary of the methods here.  It reads like a repetition of the intro.}

The current SOTA for closed-set noisy label approaches are SELF~\cite{SELF} and DivideMix~\cite{DivideMix} -- both consisting of methods that combine several of the approaches described above.
SELF~\cite{SELF} combines model ensembling, re-labelling, noisy sample identification, and data augmentation; while DivideMix~\cite{DivideMix} uses multiple model training, 
noisy sample identification, and data augmentation~\cite{MixMatch}.
These two approaches are likely to be vulnerable to open-set noise since they assume that training samples \emph{must} belong to one of the training classes -- an assumption that is not correct for open-set noise.

\section{Method}
\label{sec:method}

\subsection{Problem Definition}
\label{sec:prob_definition}

We define the training set as $\mathcal{D}=\{(\mathbf{x}_i,\mathbf{y}_i)\}_{i=1}^{|\mathcal{D}|}$, where the RGB image $\mathbf{x}:\Omega \rightarrow \mathbb{R}^3$ ($\Omega$ represents the image lattice),
the set of training labels is denoted by $\mathcal{Y}$, which forms the standard basis of $|\mathcal{Y}|$ dimensions, $\mathbf{y} \in \{0,1\}^{|\mathcal{Y}|}$ ($\sum_{c=1}^{|\mathcal{Y}|}\mathbf{y}(c)=1$, representing a multi-class problem).  Note that $\mathbf{y}_i$ is the noisy label for $\mathbf{x}_i$ and the hidden clean label is represented by $\mathbf{y}^*_i$.  

For the \textbf{closed-set noise problem with noise rate $\zeta \in [0,1]$}, we assume that $(\mathbf{x}_i,\mathbf{y}_i) \in \mathcal{D}$ is labelled as $\mathbf{y}_i = \mathbf{y}^*_i$ with probability $1-\zeta$, and $\mathbf{y}_i \sim r(\mathcal{Y})$ with probability $\zeta$, with $r(\mathcal{Y},\theta_r)$ representing a random function that picks one of the labels in $\mathcal{Y}$ following a particular distribution parameterised by $\theta_r$.

% removed \big from \big \cap
For the \textbf{open-set noise problem with noise rate $\eta \in [0,1]$}, we need to define a new training set $\mathcal{D}'$ (with $\mathcal{D}' \cap \mathcal{D} = \emptyset$), where the label set for $\mathcal{D}'$ is represented by $\mathcal{Y}'$ (with $\mathcal{Y}' \cap \mathcal{Y} = \emptyset$) -- this means that the images in $\mathcal{D}'$ no longer have labels in $\mathcal{Y}$. In such open-set problem, a proportion of $1-\eta$ samples is drawn with $(\mathbf{x}_i,\mathbf{y}_i) \in \mathcal{D}$ with $\mathbf{y}_i=\mathbf{y}^*_i$, while a proportion $\eta$ of samples are obtained with $(\mathbf{x}_i,\mathbf{y}_i) \in \mathcal{D}'$ with $\mathbf{y}_i \sim r(\mathcal{Y},\theta_r)$.

The \textbf{combined closed-set and open-set problem with rates $\rho,\omega \in [0,1]$} is defined by mixing the two types of noise above.  More specifically, $1-\rho$ of the training set contains images $(\mathbf{x}_i,\mathbf{y}_i) \in \mathcal{D}$ annotated with $\mathbf{y}_i = \mathbf{y}^*_i$, while $\omega \times \rho$ images are sampled as $(\mathbf{x}_i,\mathbf{y}_i) \in \mathcal{D}$ with label  $\mathbf{y}_i \sim  r(\mathcal{Y},\theta_r)$, and $(1 - \omega) \times \rho$ images belong to $\mathcal{D}'$ and labelled with $\mathbf{y}_i \sim r(\mathcal{Y},\theta_r)$.

% \vb{At, first, I would expect a sub-section to shortly present DivideMix. Then, a short presentation of the subjective logic.}
% \gustavo{I did that intentionally to highlight our contribution.  The risk of presenting DM and SL first is that the reader may ask -- what is the contribution here? So I decided to present the contribution first and fill the gaps later.}

\begin{figure}[t]
\centering
  \includegraphics[width=0.5\linewidth]{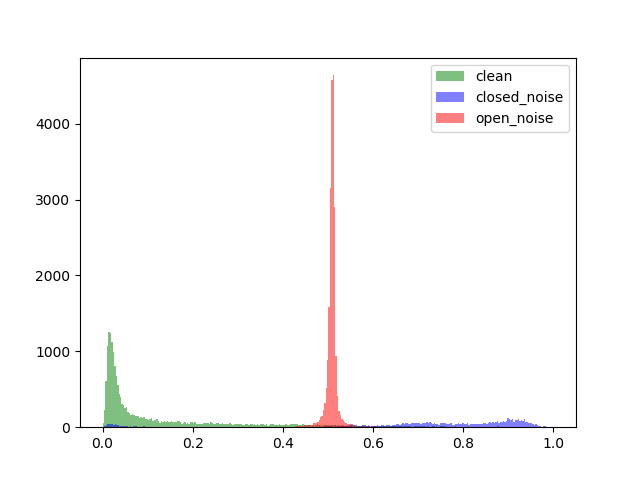}
  \caption{Per-sample loss distribution using subjective logic loss (Eq.~\eqref{eq:S_loss_sample}) with total noise ratio $\rho=0.6$ and closed-set noise rate $\omega=0.25$ (i.e. open-set noise rate is $1-\omega=0.75$) after WarmUp, with $\mathcal{D}$ from CIFAR10~\cite{krizhevsky2009learning} and $\mathcal{D}'$ from Imagenet32~\cite{chrabaszcz2017downsampled}.}
  \label{fig:SL_lossdist}
\end{figure}

\subsection{Noise Classification}
The main impediment when dealing with this problem is the need to identify closed-set and open-set noisy samples since they must be dealt differently by the method.  
One possible way of dealing with this problem is by associating closed-set samples with high losses computed from confident but incorrect classification~\cite{DivideMix}, and open-set samples with uncertain classification.  To achieve this, we propose the use of the subjective logic (SL) loss  function~\cite{sl} that relies on the theory of evidential reasoning and SL to quantify classification uncertainty.
The SL loss makes use of the Dirichlet distribution to represent subjective opinions, encoding belief and uncertainty. A network trained with the SL loss tries to learn parameters of a predictive posterior as a Dirichlet density function for the classification of the training samples. The resulting output for a given sample is considered as the evidence for the classification of that sample over a set of class labels \cite{UncertAwareAI}. Figure~\ref{fig:SL_lossdist} shows the per-sample loss distribution of training samples from a network trained using the SL loss. The separation between clean, closed-set and open-set samples is easy to capture.

\subsection{EvidentialMix}
\label{sec:edm}

Our proposed EvidentialMix simultaneously trains two networks: NetS, which uses the SL loss~\cite{sl}, and NetD, which uses the SSL training mechanism and the DivideMix (DM) loss~\cite{DivideMix}.
Broadly speaking, the ability of the SL loss to estimate classification uncertainty allows NetS to divide the training set into clean-set, open-set and closed-set samples.
The predicted clean-set and closed-set samples are then used to train NetD using MixMatch as outlined in~\cite{DivideMix}, while the predicted open-set samples are discarded for that epoch.  
Following this, NetD re-labels the \textit{entire} training data set (including predicted open-set samples) that are then used to train NetS.

As NetS iteratively learns from the labels predicted by NetD, it gets better at splitting the data into the three sets. This is so because the labels from NetD become more accurate over the training process given that it is only trained on clean and closed-set samples, and never on predicted open-set samples.
The two networks thus complement each other to produce accurate classification results for the combined closed-set and open-set noise problem.
A detailed explanation is outlined below, while Alg.~\ref{alg:EDM} delineates the full training algorithm.

\begin{algorithm}[h!]
\SetAlgoLined
\KwInput{$\mathcal{D}=\{(\mathbf{x}_i,\mathbf{y}_i)\}_{i=1}^{|\mathcal{D}|}$, number of augmentations $M$, temperature sharpening $T$, loss weights $\lambda^{(\mathcal{U})}$ and $\lambda^{(reg)}$, MixMatch parameter $\alpha$, number of epochs $E$.}
$f_{\theta^{(D)}}(c|\mathbf{x}), f_{\theta^{(S)}}(c|\mathbf{x})$ = WarmUp($\mathcal{D}$)
 
 \While{$e < E$}{
  %$\mathcal{W}^{(S)}$ = GMM($\mathcal{D}, f_{\theta^{(D)}}(c|\mathbf{x})$) \tcp{sample loss from $\theta^{(D)}$}\\
  $\mathcal{W},\mathcal{W}^{\text{op}},\mathcal{W}^{\text{cl}}$ = GMM($\mathcal{D}, f_{\theta^{(S)}}(c|\mathbf{x})$) \\
  \tcp{Train NetD}
  $\mathcal{X} = \{ (\mathbf{x}_i, \mathbf{y}_i, w_i) | (\mathbf{x}_i, \mathbf{y}_i, w_i) \in (\mathcal{D},\mathcal{W}) , w_i >   \max(w_i^{\text{op}},w_i^{\text{cl}})\}$ \\  
  $\mathcal{U} = \{\mathbf{x}_i| (\mathbf{x}_i, \mathbf{y}_i) \in \mathcal{D} , w_i^{\text{cl}} >   
  \max(w_i,w_i^{\text{op}})\}$ \\
  \For{iter=1 \textbf{to} num\_iters}   
        {
            $\{(\mathbf{x}_b, \mathbf{y}_b, w_b)\}_{b=1}^{B} \subset \mathcal{X}$ \tcp{randomly pick $B$ samples from $\mathcal{X}$}
            
            $\{\mathbf{u}_b\}_{b=1}^{B} \subset \mathcal{U}$ \tcp{randomly pick $B$ samples from $\mathcal{U}$}
            
            \For{b=1 \textbf{to} B}   
            {
                \For{m=1 \textbf{to} M}   
                {
                    $\hat{\mathbf{x}}_{b,m}$ = DataAugment($\mathbf{x}_b$)  \\
                    $\hat{\mathbf{u}}_{b,m}$ = DataAugment($\mathbf{u}_b$) \\
                }
                \For{c=1 \textbf{to} $|\mathcal{Y}|$}
                {
                    $\mathbf{p}_b(c) = \frac{1}{M}\sum_m p_{\theta^{(D)}}(c|\hat{\mathbf{x}}_{b,m})$\\
                    $\mathbf{q}_b(c) = \frac{1}{M}\sum_m p_{\theta^{(D)}}(c | \hat{\mathbf{u}}_{b,m})$ \\
                }
                $\hat{\mathbf{y}}_b$ = TempSharpen$_{T}(w_b \mathbf{y}_b + (1-w_b)\mathbf{p}_b)$ \\
                $\hat{\mathbf{q}}_b$ = TempSharpen$_{T}(\mathbf{q}_b)$ \\
            }
            $\hat{\mathcal{X}} = \{(\hat{\mathbf{x}}_{b,m},\hat{\mathbf{y}}_b)\}_{b \in (1,...,B), m \in (1,...,M)}$ \\
            $\hat{\mathcal{U}} = \{(\hat{\mathbf{u}}_{b,m},\hat{\mathbf{q}}_b) \}_{b \in (1,...,B), m \in (1,...,M)}$ \\
            $\mathcal{X}', \mathcal{U}' = \text{MixMatch}_{\alpha}(\hat{\mathcal{X}},\hat{\mathcal{U}}$) \\
            $\theta^{(D)} = \text{SGD}(\mathcal{L}^{(D)},\theta^{(D)},\mathcal{X}', \mathcal{U}')$ \\
        }
        \tcp{Train NetS}
        \For{i=1 \textbf{to} $|\mathcal{D}|$}
        {
            $\hat{c}_i = \arg\max_{c \in \mathcal{Y}} \left [ (w_i^{\text{cl}}) p_{\theta^{(D)}}(c|\mathbf{x}_i) +
                (1-w_i^{\text{cl}}) \mathbf{y}_i(c)) \right ]$\\
            $\hat{\mathbf{y}}_i = \text{onehot}(\hat{c}_i) $\\
        }
        $\theta^{(S)} = SGD(\mathcal{L}^{(S)},\theta^{(S)},\{ (\mathbf{x}_i,\hat{\mathbf{y}}_i) \}_{i=1}^{|\mathcal{D}|}$) \\
 }
 \label{alg:EDM}
 \caption{EvidentialMix (EDM)}
\end{algorithm}
 %\caption{EvidentialMix (EDM)}

Algorithm~\ref{alg:EDM} trains NetD, represented by $f_{\theta^{(D)}}(c|\mathbf{x})$, and NetS, denoted by $f_{\theta^{(S)}}(c|\mathbf{x})$ (with $c \in \{1,...,|\mathcal{Y}|\}$) -- both of which return a logit in $\mathbb R^{|\mathcal{Y}|}$.  In the warm-up stage (see line WarmUp($\mathcal{D}$)), we train both models for a limited number of epochs using the cross entropy loss for $f_{\theta^{(D)}}(c|\mathbf{x})$, where probability is obtained by applying a softmax activation $\sigma(.)$ to form $p_{\theta^{(D)}}(c|\mathbf{x}) = \sigma(f_{\theta^{(D)}}(c|\mathbf{x}))$, and the following (SL) loss for $f_{\theta^{(S)}}(c|\mathbf{x})$~\cite{sl}:
\begin{equation}
    \mathcal{L}^{(S)} = \frac{1}{|\mathcal{D}|}\sum_{i=1}^{|\mathcal{D}|} \ell^{(S)}(\mathbf{x}_i,\mathbf{y}_i,\theta^{(S)}),
    \label{eq:S_loss}
\end{equation}
with~\cite{sl}:
\begin{equation}
    \ell^{(S)}(\mathbf{x}_i,\mathbf{y}_i,\theta^{(S)}) = \sum_{c = 1}^{|\mathcal{Y}|} (\mathbf{y}_i(c) - \alpha_{ic}/S_i)^2 + \frac{\alpha_{ic} (S_i - \alpha_{ic})}{S_i^2(S_i+1)},
    \label{eq:S_loss_sample}
\end{equation}
where $\alpha_{ic}=\varphi(f_{\theta^{(S)}}( c | \mathbf{x}_i)) + 1$ for class $c \in \{1,...,|\mathcal{Y}|\}$, with $\varphi(.)$ representing the ReLU activation function, and 
$S_i = \sum_{c = 1}^{|\mathcal{Y}|} \alpha_{ic}$.

The classification of samples into the clean, closed-set, and open-set is performed using the SL loss values from Eq.~\eqref{eq:S_loss_sample} for the entire training set $\mathcal{D}$.  More specifically, we take the set of losses $\{ \ell^{(S)}(\mathbf{x}_i,\mathbf{y}_i,\theta^{(S)}) \}_{i=1}^{|\mathcal{D}|}$ and fit a $\psi$-component Gaussian mixture model (GMM) using the Expectation-Maximization algorithm.  The idea we explore in this paper lies in the fact that the model output for the \textbf{clean samples} will tend to be \emph{confident and at the same time agree with its original label, producing a small loss}.  
The model output for \textbf{closed-set noise samples} will also tend to be \emph{confident but at the same time disagree with its original label, generating a large loss value}.
The model output for \textbf{open-set noise samples}, however, will \emph{not produce a confident output, resulting in a loss value that is neither large nor small}.
Therefore, the multi-component GMM will capture each of these sets, where the clean probability $w_i$ is the posterior probability $p(\mathcal{G}|\ell^{(S)}(\mathbf{x}_i,\mathbf{y}_i,\theta^{(S)}))$, where $\mathcal{G}$ denotes the set of Gaussian components with mean $\leq \mu_{min}$ (i.e., small losses), the closed-set posterior probability $w_i^{\text{cl}}=p(\mathcal{G}^{\text{cl}}|\ell^{(S)}(\mathbf{x}_i,\mathbf{y}_i,\theta^{(S)}))$ is computed from the components with mean $\geq \mu_{max}$ (i.e., large losses), and the open-set probability $w_i^{\text{op}}$ is the posterior of the remaining components with means $\in (\mu_{min},\mu_{max})$ -- these posterior probabilities form the three sets $\mathcal{W}=\{ w_i \}_{i=1}^{|\mathcal{D}|}$, $\mathcal{W}^{\text{cl}}=\{ w^{\text{cl}}_i \}_{i=1}^{|\mathcal{D}|}$ and $\mathcal{W}^{\text{op}}=\{ w^{\text{op}}_i \}_{i=1}^{|\mathcal{D}|}$.  Using these posteriors, we can build the set of clean samples, represented by $\mathcal{X}$ with samples where the clean posterior probability $w_i$ is larger than the other two probabilities, and the closed-set, denoted by $\mathcal{U}$, containing samples that have the closed-set posterior probability $w_i^{\text{cl}}$ larger than the other two.

%\gustavo{need a figure showing this 3-class GMM classification, maybe like the one in Fig.~\ref{fig:S_lossdist}.}

% Sensoy et al. proposed this loss function with the aim of extending classical deep learning with the ideas from evidential reasoning and subjective logic to quantify uncertainty in classification tasks \cite{sl}. Subjective logic uses Dirichlet distributions to represent subjective opinions, which encodes belief and uncertainty. For a sample, a network trained using SL  loss tries to learn parameters of a predictive posterior as a Dirichlet density function for the classification of the sample. This is done by replacing the softmax layer in a classical deep classifier with an activation function that produces only non-negative outputs (e.g., ReLU). The resulting output is considered as the evidence for the classification of the sample over $C$ class labels. From the calculated evidence, parameters of the corresponding Dirichlet density is calculated as $\alpha$ = evidence + 1. Hence, when the total evidence is zero, the resulting Dirichlet density corresponds to the uniform distribution, i.e., a Dirichlet density whose all parameters are one. It is mapped to a subjective opinion with zero belief and total uncertainty \cite{UncertAwareAI}.

Next, we train NetD with the clean set  $\mathcal{X}$ and closed-set $\mathcal{U}$, defined above.
A mini-batch is sampled from $\mathcal{X}$ and $\mathcal{U}$, and we augment each sample in each set $M$ times~\cite{DivideMix}.  
The average classification probabilities for the clean and closed-set samples are then computed using the $M$ augmented samples, which after temperature sharpening, denoted by TempSharpen$_T(.)$ with $T$ denoting the temperature, form the `new' samples and labels for the clean and closed-set samples, $\hat{\mathcal{X}} = \{ (\hat{\mathbf{x}}_{b,m},\hat{\mathbf{y}}_b)\}_{b,m=1}^{B,M}$ and 
$\hat{\mathcal{U}} = \{ (\hat{\mathbf{u}}_{b,m},\hat{\mathbf{q}}_b)\}_{b,m=1}^{B,M}$, respectively.
The last stage before stochastic gradient descent (SGD) is the MixMatch process~\cite{MixMatch}, where samples from the $\hat{\mathcal{X}}$ and $\hat{\mathcal{U}}$ are linearly combined to form $\mathcal{X}'$ and $\mathcal{U}'$.
SGD minimises the DM loss that combines the following functions~\cite{DivideMix}:
\begin{equation}
    \mathcal{L}^{(D)} = \mathcal{L}^{(\mathcal{X})} + \lambda^{(\mathcal{U})}\mathcal{L}^{(\mathcal{U})} + \lambda^{(reg)}\mathcal{L}^{(reg)},
    \label{eq:D_loss_full}
\end{equation}
where $\lambda^{(\mathcal{U})}$ denotes the weight of the loss associated with the unlabelled data set and $\lambda^{(reg)}$ weights the regularisation loss.
The loss terms in Eq.~\eqref{eq:D_loss_full} are defined by
\begin{equation}
    \mathcal{L}^{(reg)} = \sum_{c=1}^{|\mathcal{Y}|} \frac{1}{|\mathcal{Y}|} \log \left ( \frac{1}{|\mathcal{X}'| + |\mathcal{U}'|} \sum_{\mathbf{x} \in (\mathcal{X}' \bigcup \mathcal{U}')} p_{\theta^{(D)}}(c|\mathbf{x}) \right ),
    \label{eq:L_reg}
\end{equation}
and
\begin{equation}
    \begin{split}
        \mathcal{L}^{(\mathcal{X})} &= -\frac{1}{|\mathcal{X}'|} \sum_{(\hat{\mathbf{x}},\hat{\mathbf{y}})\in \mathcal{X}'} \sum_{c=1}^{|\mathcal{Y}|} \hat{\mathbf{y}}(c) \log(p_{\theta^{(D)}}(c | \hat{\mathbf{x}}) ), \\
        \mathcal{L}^{(\mathcal{U})} &= \frac{1}{|\mathcal{U}'|} \sum_{(\hat{\mathbf{u}},\hat{\mathbf{q}}) \in \mathcal{U}'} \| \hat{\mathbf{q}} - p_{\theta^{(D)}}(: | \hat{\mathbf{u}}) \|_2^2,
    \end{split}
    \label{eq:L_X_L_U}
\end{equation}
where $p_{\theta^{(D)}}(: | \hat{\mathbf{u}}) \in [0,1]^{|\mathcal{Y}|}$ represents the model output of all $|\mathcal{Y}|$ labels for input $\hat{\mathbf{u}}$.
After training NetD, we train the NetS model $f_{\theta^{(S)}}(c|\mathbf{x})$ by minimising the SL loss Eq.~\eqref{eq:S_loss} with an updated training set, represented by $\{ (\mathbf{x}_i,\hat{\mathbf{y}}_i)\}_{i=1}^{|\mathcal{D}|}$, formed by  $\hat{c}_i = \arg\max_c \left [ (w_i^{\text{cl}}) p_{\theta^{(D)}}(c|\mathbf{x}_i) +
                (1-w_i^{\text{cl}}) \mathbf{y}_i(c)) \right ]$, for $i \in \{1,...,|\mathcal{D}|\}$
that produces the new labels $\hat{\mathbf{y}}_i = \text{onehot}(\hat{c}_i)$.  

The inference for a test sample $\mathbf{x}$ relies entirely on the NetD classifier, as follows: $c^* = \arg\max_{c \in \mathcal{Y}} p_{\theta^{(D)}}(c|\mathbf{x})$.  

\subsection{Implementation}
\label{sec:implementation}

We train an 18-layer PreAct Resnet~\cite{he2016identity} (for both NetS and NetD) using stochastic gradient descent (SGD) with momentum of 0.8, wight decay of 0.0005 and batch size of 64. 
The learning rate is 0.02 for WarmUp and for 100 epochs in the main training process, which is reduced to 0.002 afterwards.
The WarmUp stage lasts for 10 and 30 epochs for NetD and NetS, respectively, where NetD is trained with a cross-entropy loss (i.e., $\mathcal{L}^{(\mathcal{X})}$ in Eq.~\eqref{eq:L_X_L_U} using the unchanged training set $\mathcal{D}$) while NetS is trained with the subjective logic loss $\mathcal{L}^{(S)}$ in Eq.~\eqref{eq:S_loss} also using $\mathcal{D}$.
After WarmUp, both models are trained for $E=200$ epochs.
Similar to~\cite{DivideMix}, the number of data augmented samples is
$M=2$, the sharpening temperature is $T=0.5$, the MixMatch parameter is $\alpha=4$, and the regularisation weight for the DM loss in Eq.~\eqref{eq:D_loss_full} is $\lambda^{(reg)}=1$. 
However, unlike ~\cite{DivideMix} that manually selects the value of $\lambda^{(\mathcal{U})}$ based on the value of $\rho$, we set $\lambda^{(\mathcal{U})}$ = 25 for all our experiments. 
For the GMM, we use $\psi=20$ components, $\mu_{min}=0.3$, and $\mu_{max}=0.7$ since these values produced stable results.  
%Therefore, under no circumstances do we change the value of any of our hyperparameters based on the noise ratio ($\rho, \omega$).

\newcolumntype{P}[1]{>{\centering\arraybackslash}p{#1}}
\begin{table*}[t]
% \small
\footnotesize
\centering
\begin{tabular}{c|l c|P{0.7cm}| P{0.7cm} |P{0.7cm}|P{0.7cm}|P{0.7cm}||P{0.7cm}|P{0.7cm}|P{0.7cm}|P{0.7cm}|P{0.7cm}}
\toprule
\multicolumn{2}{c}{}  & $\rho$ &  \multicolumn{5}{c}{0.3}& \multicolumn{5}{c}{0.6}  \\ 
\cmidrule{3-13}
\multicolumn{2}{c}{}& $\omega$ & 0 & 0.25 & 0.5  & 0.75 & 1 & 0 & 0.25 & 0.5 & 0.75 & 1 \\
\midrule
\multirow{8}{*}{\rotatebox[origin=c]{90}{ImageNet32}}  & \multirow{2}{*}{RoG~\cite{lee2019robust}}   & Best    & 91.9 & 90.7 & 90.2 & 89.6 & 89.5 & 87.8 &85.7 &84.5 &83.1 & 82.9 \\
                                                    &                           & Last    & 91.0 & 88.7 & 86.6 & 86.2 & 83.9 & 85.9 &78.1 &70.3 &64.7 & 59.8 \\
\cmidrule{2-13}
                                                    &\multirow{2}{*}{ILON~\cite{OpenSetPaper}}      & Best    &91.8  &90.7 & 88.0 & 86.5 & 85.8 & 87.7 & 83.4 &81.2 &78.7 &77.3 \\
                                                    &                           & Last    &90.6  &86.9 &82.0 & 77.3 &72.7 &85.5 &72.6 &58.9 &54.4 & 46.5 \\
\cmidrule{2-13}
                                                    & \multirow{2}{*}{DivideMix~\cite{DivideMix}}  & Best    & 92.4 & 92.5 & 93.4 & 93.9 & 94.3 & \textbf{92.5} & 92.8 & 93.2 &93.9 & \textbf{94.7} \\
                                                    &                            & Last    & 92.0 & 92.5 & 93.0 & 93.7 & 94.1 & \textbf{92.5} & 92.2 & 92.8 & 93.2 & \textbf{94.6} \\
\cmidrule{2-13}
                                                    &  \multirow{2}{*}{\textbf{EDM (Ours)}}        & Best    & \textbf{93.2}  & \textbf{94.4} & \textbf{94.7} & \textbf{95.1} & \textbf{95.2} & 91.2 & \textbf{93.7} & \textbf{94.0} & \textbf{94.1} & 94.1 \\
                                                    &                           & Last    & \textbf{92.5} & \textbf{93.7} & \textbf{94.5} & \textbf{94.7} & \textbf{94.8} & 90.9 & \textbf{93.1} & \textbf{93.4} & \textbf{93.9} & 94.1 \\
\midrule
%%%middle
\multirow{8}{*}{\rotatebox[origin=c]{90}{CIFAR-100}}& \multirow{2}{*}{RoG~\cite{lee2019robust}} & Best   & 91.4             & 90.9             & 89.8
                                                    &90.4              &89.9             &88.2              &85.2              &84.1              & 83.7 & 83.1  \\
                                                    &                      & Last   & 89.8             & 87.4             & 85.9             & 84.9
                                                    & 84.5             &82.1              &72.9             &66.3            &62.0            &59.5  \\
\cmidrule{2-13}
                                                    & \multirow{2}{*}{ILON~\cite{OpenSetPaper}}& Best    & 90.4& 88.7 & 87.4 & 87.2 &86.3 &83.4 &82.6 &80.5 &78.4 & 77.1 \\
                                                    &                               & Last    & 87.4 & 84.3 & 80.0 & 74.6 &73.8 &78.0 &67.9 &55.2 &48.7 & 45.6 \\
\cmidrule{2-13}
                                                    &\multirow{2}{*}{DivideMix~\cite{DivideMix}} & Best    & 89.3 & 90.5 & 91.5 & 93.0 & 94.3 & 89.0 &90.6 & 91.8 & 93.4 & \textbf{94.4} \\
                                                    &                           & Last    & 88.7 & 90.1 & 90.9 & 92.8 & 94.0 & 88.7 & 89.8 & 91.5 & 93.0 & \textbf{94.3} \\
\cmidrule{2-13}
                                                    & \multirow{2}{*}{\textbf{EDM (Ours)}}                      & Best    & \textbf{92.9}  & \textbf{93.8} & \textbf{94.5} & \textbf{94.8} &\textbf{95.3} &\textbf{90.6} & \textbf{92.9} &\textbf{93.4} &\textbf{93.7} & 94.3 \\
                                                    &                    & Last    & \textbf{91.9} & \textbf{93.1} & \textbf{94.0} & \textbf{94.5} &\textbf{95.1} &\textbf{89.4} &\textbf{91.4} &\textbf{92.8} &\textbf{93.4} & 94.0 \\
 
 \bottomrule
\multicolumn{12}{c}{}
\end{tabular}
\caption{Benchmark results of all competing methods and our proposed EDM. Clean data was sampled from CIFAR10, while the open-set noise came from ImageNet32 and CIFAR-100. The total noise in the training data is represented by $\rho \in \{0.3,0.6\}$, where the closed-set proportion of this noise is $\omega \in \{0,0.25,0.5,0.75,1\}$ and open-set proportion is $1-\omega$.\label{tab:results}}
\end{table*}

\begin{figure*}[t]
\centering
\hspace*{0mm}{\scriptsize \hspace{5mm} ImageNet32 \hspace{45mm} CIFAR-100}  \\
\noindent \hspace*{5mm}  \rule{5cm}{0.5pt} \hspace{7mm} \rule{5cm}{0.5pt}
\\
% \hspace*{9mm}{\scriptsize \hspace{1mm}\asmall \hspace{13mm} \amedium \hspace{15mm} \alarge \hspace{14mm} \fsmall \hspace{2.5mm}}\\
\begin{subfigure}{.18\textwidth}
  \centering
  % include first image
  \includegraphics[width=\linewidth]{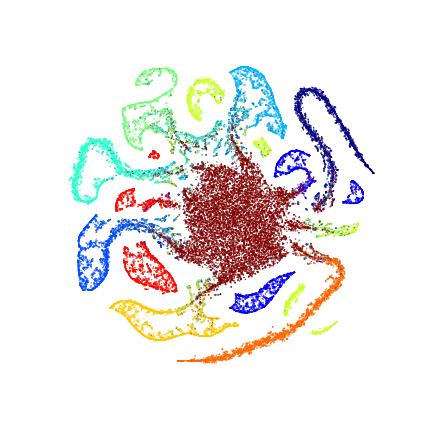}
  \caption*{RoG~\cite{lee2019robust}}
\end{subfigure}
\begin{subfigure}{.18\textwidth}
  \centering
  % include second image
  \includegraphics[width=\linewidth]{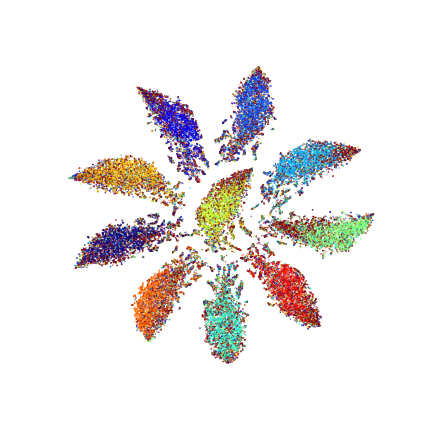}
  \caption*{ILON~\cite{OpenSetPaper}}
\end{subfigure}
\begin{subfigure}{.18\textwidth}
  \centering
  % include second image
 \includegraphics[width=\linewidth]{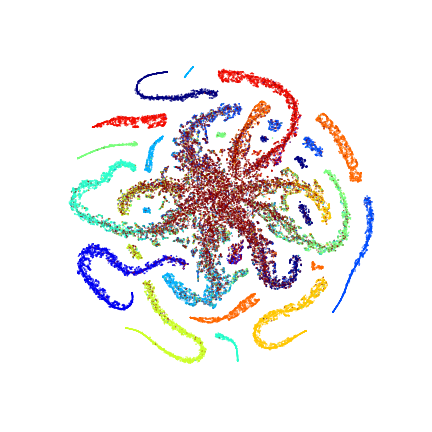}
 \caption*{RoG~\cite{lee2019robust}}
\end{subfigure}
\begin{subfigure}{.18\textwidth}
  \centering
  % include second image
  \includegraphics[width=\linewidth]{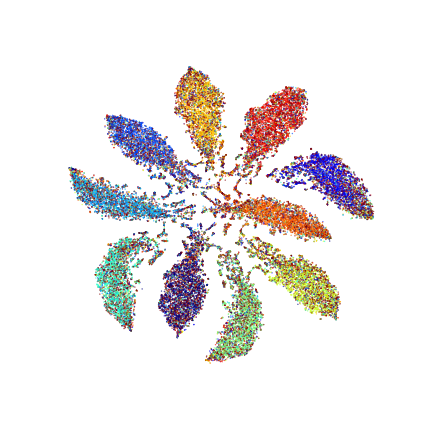}
  \caption*{ILON~\cite{OpenSetPaper}}
\end{subfigure}
\\
\begin{subfigure}{.18\textwidth}
  \centering
  % include first image
  \includegraphics[width=\linewidth]{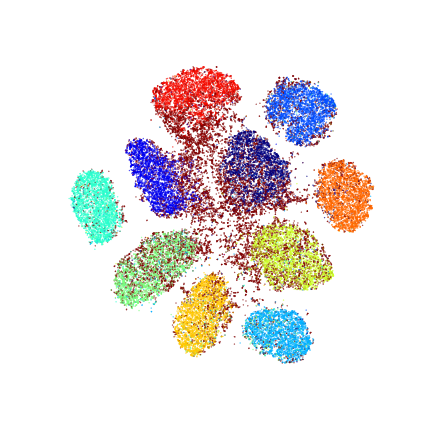} 
  \caption*{DivideMix~\cite{DivideMix}}
\end{subfigure}
\begin{subfigure}{.18\textwidth}
  \centering
  % include second image
  \includegraphics[width=\linewidth]{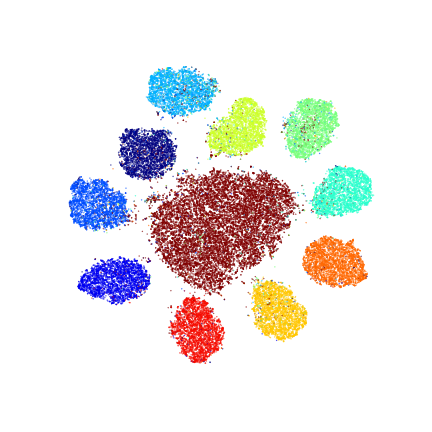}
  \caption*{EDM [ours]}
\end{subfigure}
\begin{subfigure}{.18\textwidth}
  \centering
  % include second image
  \includegraphics[width=\linewidth]{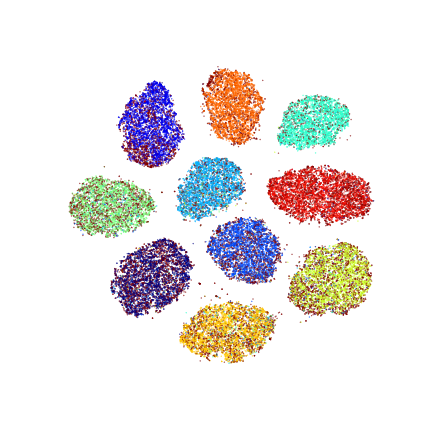}
  \caption*{DivideMix~\cite{DivideMix}}
\end{subfigure}
\begin{subfigure}{.18\textwidth}
  \centering
  % include second image
  \includegraphics[width=\linewidth]{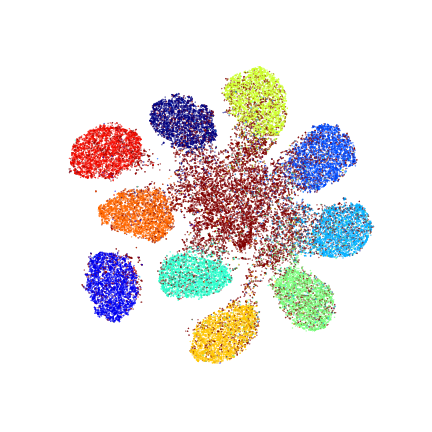}
  \caption*{EDM [ours]}
\end{subfigure}
% \begin{subfigure}{\textwidth}
%   \centering
%   % include second image
%   \includegraphics[width=0.5\linewidth]{images/loss_dist/legend.pdf} 
% \end{subfigure}
\caption{t-SNE plots of the related methods and our proposed EDM, where the total noise rate is $\rho=0.6$ with the closed-set proportion being $\omega = 0.5$, and CIFAR-100 and ImageNet32 representing open-set data sets.  The \textbf{brown} samples represent the open-set noise, while the other colours denote the true CIFAR-10 classes.}
\label{fig:tsne}
\end{figure*}

\begin{figure*}[t]
\centering
\hspace*{8mm}{\scriptsize EDM [ours] \hspace{50mm} DivideMix \cite{DivideMix}}  \\
\noindent  \rule{5cm}{0.5pt} \hspace{12mm} \rule{5cm}{0.5pt}
\\

\begin{subfigure}{.18\textwidth}
  \centering
  % include first image
  \includegraphics[width=\linewidth]{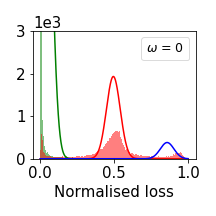} 
\end{subfigure}
\begin{subfigure}{.18\textwidth}
  \centering
  % include second image
  \includegraphics[width=\linewidth]{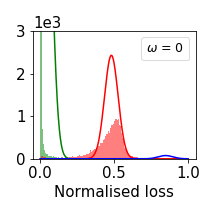} 
\end{subfigure}
\begin{subfigure}{.18\textwidth}
  \centering
  % include second image
  \includegraphics[width=\linewidth]{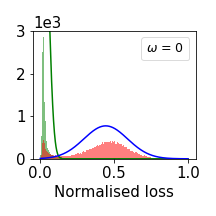} 
\end{subfigure}
\begin{subfigure}{.18\textwidth}
  \centering
  % include second image
  \includegraphics[width=\linewidth]{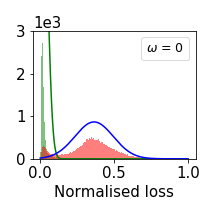} 
\end{subfigure}
\\
\begin{subfigure}{.18\textwidth}
  \centering
  % include first image
  \includegraphics[width=\linewidth]{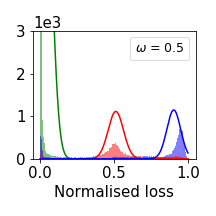} 
\end{subfigure}
\begin{subfigure}{.18\textwidth}
  \centering
  % include second image
  \includegraphics[width=\linewidth]{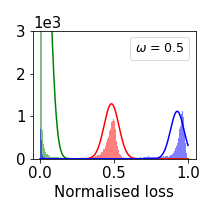} 
\end{subfigure}
\begin{subfigure}{.18\textwidth}
  \centering
  % include second image
  \includegraphics[width=\linewidth]{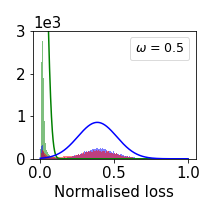} 
\end{subfigure}
\begin{subfigure}{.18\textwidth}
  \centering
  % include second image
  \includegraphics[width=\linewidth]{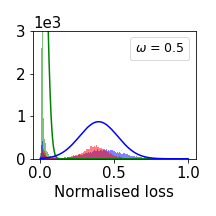} 
\end{subfigure}
\\
\begin{subfigure}{.18\textwidth}
  \centering
  % include first image
  \includegraphics[width=\linewidth]{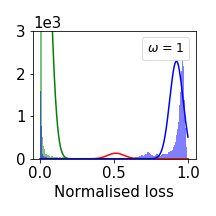} 
\end{subfigure}
\begin{subfigure}{.18\textwidth}
  \centering
  % include second image
  \includegraphics[width=\linewidth]{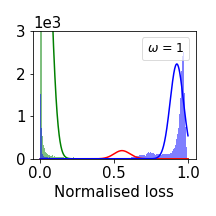} 
\end{subfigure}
\begin{subfigure}{.18\textwidth}
  \centering
  % include second image
  \includegraphics[width=\linewidth]{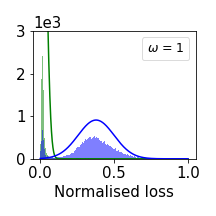} 
\end{subfigure}
\begin{subfigure}{.18\textwidth}
  \centering
  % include second image
  \includegraphics[width=\linewidth]{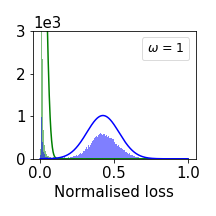} 
\end{subfigure}
\\
\begin{subfigure}{\textwidth}
  \centering
  % include second image
  \includegraphics[width=0.4\linewidth]{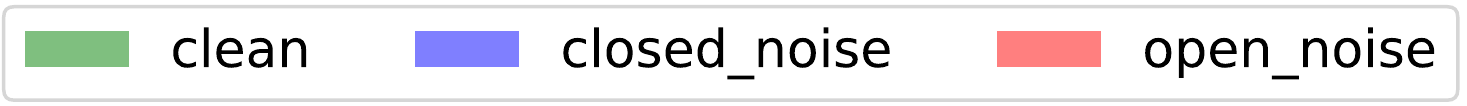} 
\end{subfigure}
\caption{Per-sample loss distributions for the open-set (orange), closed set (blue) and clean set (green) samples produced by EDM (left) and DivideMix (right) at epoch $e=100$, where $\rho=0.6$ and $\omega \in \{0,0.5,1\}$, with open-set datasets $\in$ \{CIFAR-100, ImageNet32\}.  We also show the estimated GMM posterior probabilities for the clean, open-set and closed-set noise samples using our EDM, and for the clean and noisy samples using DivideMix~\cite{DivideMix}.}
\label{fig:EDMvsDM}
\end{figure*}

%\hspace*{3mm}{\scriptsize \asmall \hspace{18mm} \amedium \hspace{20mm} \alarge \hspace{18mm} \fsmall}\\

\section{Experiments}
\label{sec:experiments}

%\gustavo{introduce datasets, CIFAR10, CIFAR100, Imagenet32}

Following the prior work on closed-set and open-set noise problems~\cite{DivideMix,OpenSetPaper,lee2019robust}, we conduct our experiments on the CIFAR-10 data set~\cite{krizhevsky2009learning} for closed-set noise~\cite{DivideMix,lee2019robust}; and include the CIFAR-100 (small-scale)~\cite{krizhevsky2009learning} and ImageNet32 (large-scale)~\cite{chrabaszcz2017downsampled} data sets for the open-set noise scenario~\cite{OpenSetPaper,lee2019robust}. CIFAR-10 has 10 classes with 5000 32$\times$32 pixel training images per class (forming a total of 50000 training images), and a testing set with 10000 32$\times$32 pixel images with 1000 images per class. CIFAR-100 has 100 classes with 5000 32$\times$32 pixel images per class and ImageNet32 is a down-sampled variant of ImageNet~\cite{deng2009imagenet}, with 1281149 images and 1000 classes, but resized to 32$\times$32 pixels per image. 
All data sets above have been set up with curated labels, so below we introduce a new noisy label benchmark evaluation that combines closed-set and open-set synthetic label noise.

%As CIFAR-10 has been built with curated labels, we introduce a combination of closed-set and open-set synthetic label noise.% to measure robustness to label noise. 
%For evaluation, we use clean CIFAR10 validation set containing 10000 32$\times$32 pixel images.
%Our setup is built with the following combination of closed-set and open-set label noise. It is imperative to mention that the samples selected to introduce closed-set and open-set noise were mutually exclusive.

\subsection{Combined Open-set and Closed-set Noisy Label Benchmark}
\label{sec:proposed_benchmark}

The proposed benchmark is defined by the rate of label noise in the experiment, denoted by $\rho \in \{0.3,0.6\}$, and the proportion of closed-set noise in the label noise, denoted by $\omega \in \{0,0.25,0.5,0.75,1\}$.  
The closed-set label noise is simulated by randomly selecting $\rho\times\omega\times100\%$ of the training samples from CIFAR-10, and symmetrically shuffling their label, similarly to the synthetic label noise used in~\cite{DivideMix}.
% , where the true label can be included when defining a random label to a sample.
The open-set label noise is simulated by randomly selecting $\rho\times(1-\omega)\times100\%$ of the training images from CIFAR-10 and replacing them with images randomly selected from either CIFAR-100~\cite{krizhevsky2009learning} or ImageNet32~\cite{chrabaszcz2017downsampled}, where a CIFAR-10 label is randomly assigned to each one of these images.
Results are based on the classification accuracy on the clean testing set from CIFAR-10, using the benchmark proposed above. 
We also show a comparison of the sample distribution between EDM and other related approaches in the feature space using t-SNE~\cite{maaten2008visualizing}, and the effectiveness of EDM to separate clean, closed-set and open-set noisy samples.
%CIFAR-100 has 100 classes with 5000 32$\times$32 pixel images per class and ImageNet32 is a downsampled variant of ImageNet~\cite{deng2009imagenet}, with exactly the same number of images and classes (1K) as ImageNet, but resized to 32$\times$32 pixels per image. 
% \gustavo{for imagenet, do we have 1000 images per class and 1000 classes?}
% \gustavo{introduce open, closed and combined noise setup and the proposed benchmark with measures}
% \gustavo{all details about our method -- model, hjyperparmeters, mini-batch size, data augmentation, all variables in Sec.~\ref{sec:method} must be defined here, or we should say we did CV to do model selection.  Learning rate, number of epochs, optimiser.}

\newcommand\hmOneOfNine[1]{\includegraphics[width=0.22\linewidth]{images/loss_dist/#1} }%
\newcommand\omegazero[0]{%
\hmOneOfNine{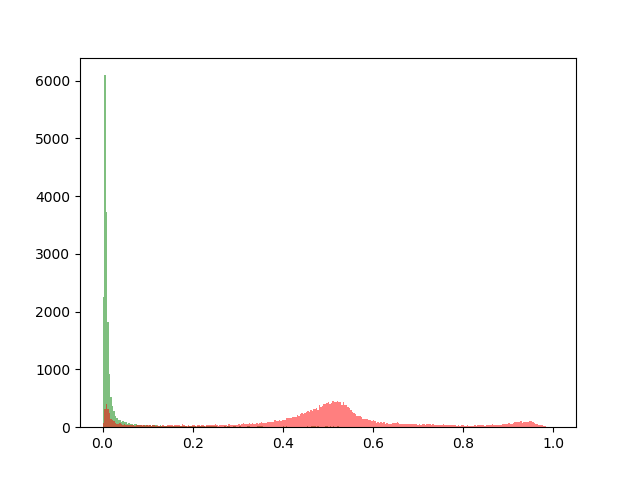}%
\hmOneOfNine{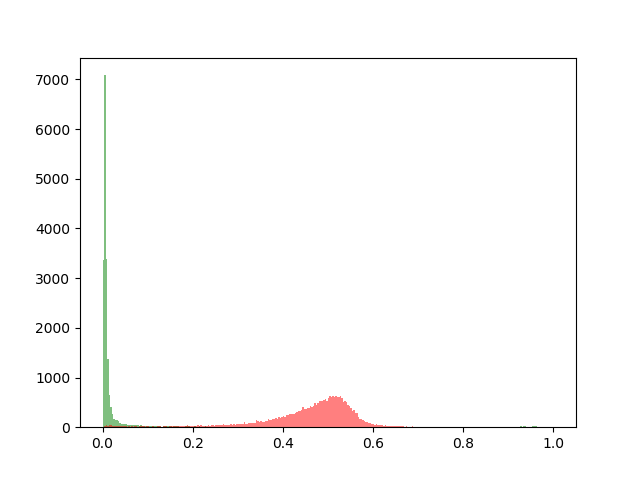}%
\hmOneOfNine{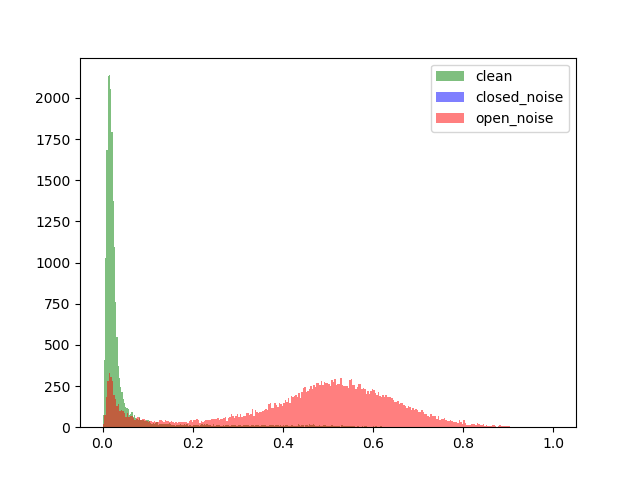}%
\hmOneOfNine{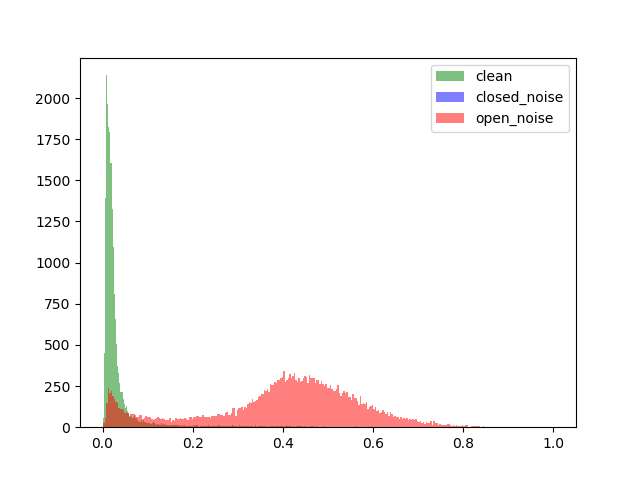}%
\vspace{-2mm}
}

\newcommand\omegahalf[0]{%
\hmOneOfNine{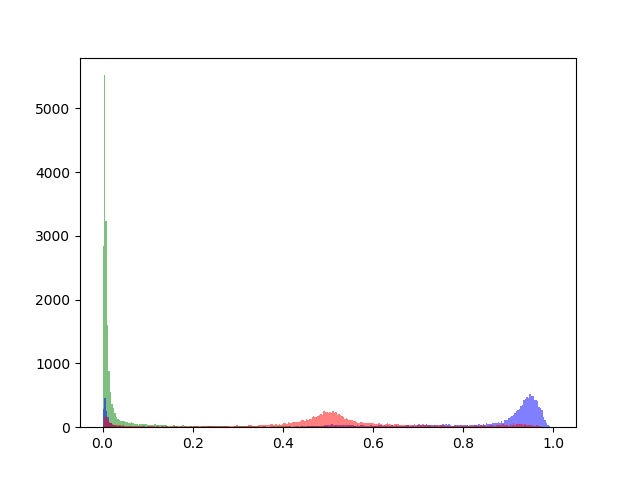}%
\hmOneOfNine{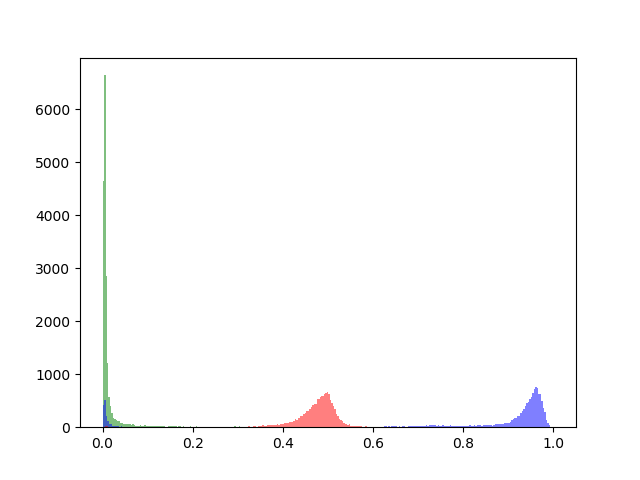}%
\hmOneOfNine{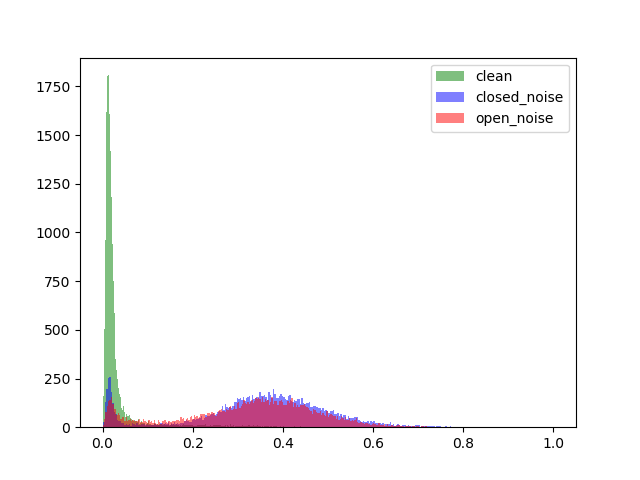}%
\hmOneOfNine{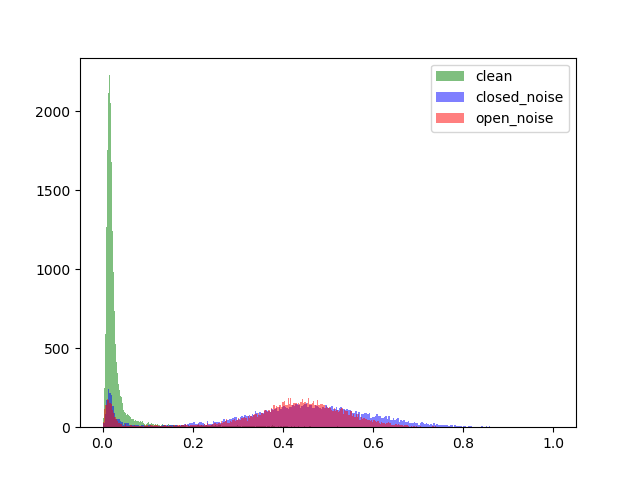}%
\vspace{-2mm}
}

\newcommand\omegaone[0]{%
\hmOneOfNine{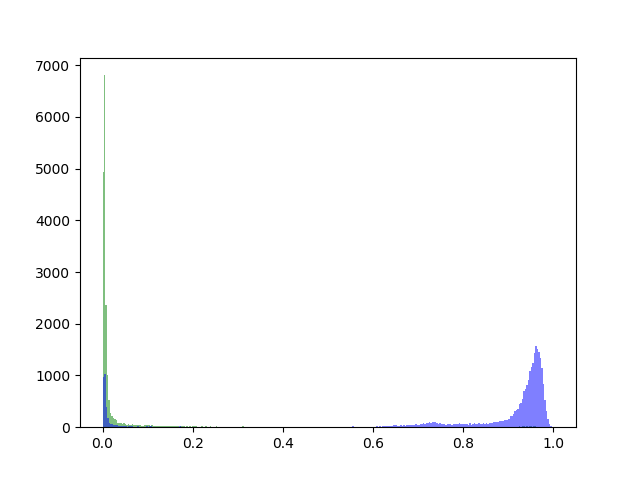}%
\hmOneOfNine{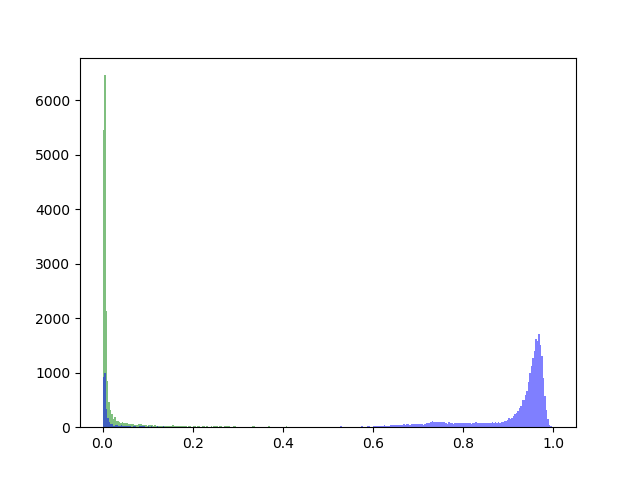}%
\hmOneOfNine{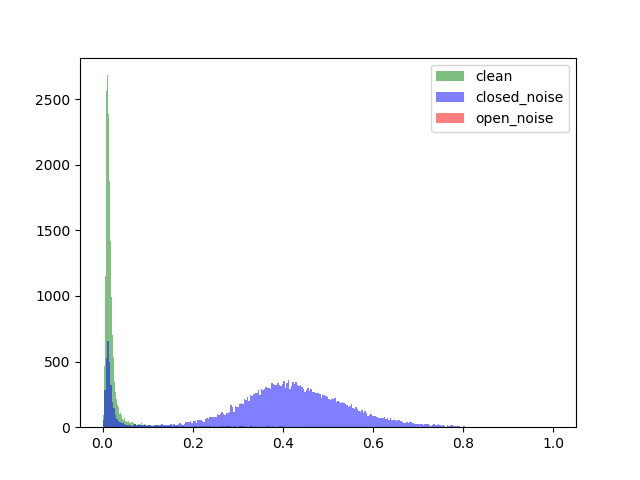}%
\hmOneOfNine{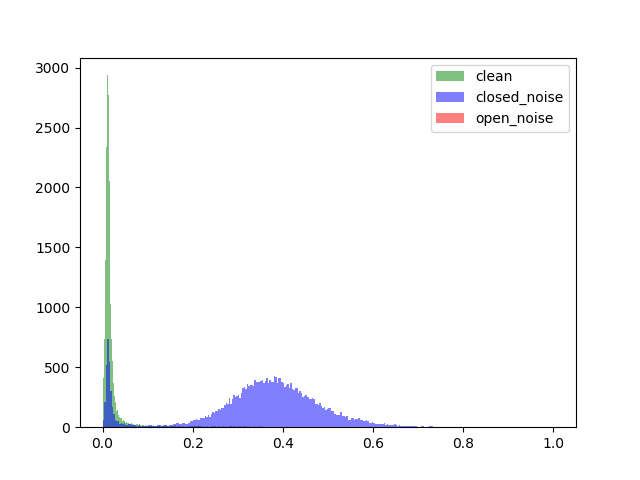}%
\vspace{-2mm}
}

\newcommand\asmall[0]{CIFAR-100\xspace}
\newcommand\amedium[0]{Imagenet32\xspace}
\newcommand\alarge[0]{CIFAR-100\xspace}
\newcommand\fsmall[0]{Imagenet32\xspace}

\subsection{Related Approaches for Comparison}
\label{sec:competing_approaches}

We compare our proposed approach with the three methods listed below:

\textbf{DivideMix~\footnote{\label{originalcode}We used the publicly available code provided by the authors of the paper to produce our results.}}~\cite{DivideMix} is the current SOTA method that converts the problem of closed-set noisy label learning into a semi-supervised learning problem. It follows a multiple-model approach that splits the training data into clean and noisy subsets by fitting a 2-component Gaussian Mixture Model (GMM) to the loss values of the training samples at each epoch. Next, the framework discards the labels of the predicted noisy samples and uses MixMatch~\cite{MixMatch} to train the model. 
%It is worth mentioning that DivideMix has not been designed to handle open-set noise and hence cannot identify open-set noise samples as evident from the plots in Fig~\ref{fig:EDMvsDM}.

\textbf{ILON~\footnote{As the authors did not make their code publicly available, we implemented their method from scratch and trained a Siamese network of 18-layer PreAct Resnet to produce our results.}}~\cite{OpenSetPaper} introduces the open-set noisy label learning problem, where the proposed approach is based on an iterative solution that re-weights the samples based on the outlier measure of the Local Outlier Factor Algorithm (LOF)~\cite{kriegel2009loop}. 
%Clean and noisy samples are identified to train a Siamese network with a contrastive loss, which pushes clean and noisy samples to be dissimilar, and a weighted softmax loss to avoid learning from noisy samples.

\textbf{RoG~}\textsuperscript{\ref{originalcode}}~\cite{lee2019robust} builds an ensemble of generative classifiers formed from features extracted from multiple layers of the the ResNet model. The authors of RoG tested their approach on both closed-set noise and open-set noise \textit{separately} which makes it an important benchmark to consider in our \textit{combined} setup.

\subsection{Results and Discussion}

\textbf{Classification accuracy:} Table~\ref{tab:results} shows the results computed from the  benchmark evaluation of the proposed EDM, in comparison with the results of RoG~\cite{lee2019robust}, ILON~\cite{OpenSetPaper}, and DivideMix~\cite{DivideMix}.  The evaluation relies on different rates of label noise ($\rho \in \{0.3,0.6\}$) and closed-set noise ($\omega\in \{0,0.25,0.5,0.75,1 \}$) using CIFAR-100 and ImageNet32 as open-set data sets. Results show that our method EDM outperforms all the competing approaches for 17 out of the 20 noise settings and is a close second on the remaining 3. For $\rho=0.3$, EDM produces better results than all competing methods for all values of $\omega$ and choice of open-set data sets, with an improvement of more than $3\%$ over the next best method in some cases. On the other hand, both RoG and ILON perform significantly worse than EDM and DivideMix, particularly for $\rho=0.6$ where the difference in accuracy is over 15\% in some cases.
In general, RoG and ILON are observed to perform worse when the proportion of closed-set noise increases, while the converse is true for DivideMix and EDM. It is also apparent that EDM is more robust to open-set noise than DivideMix, as evident from classification results when $\omega$ is small.

% \gustavo{Ragav -- please check if that is how you make the tsne graphs.}
\textbf{Feature representations:} We show the t-SNE plots~\cite{maaten2008visualizing} in Fig.~\ref{fig:tsne} for all the methods for the case where total noise rate $\rho=0.6$, with the closed-set proportion $\omega =0.5$ using CIFAR-100 and ImageNet32 as open-set data sets.  
In particular, the features for all methods are extracted from the last layer of the models (in our case, we use the features from the NetD, which is the one used for classification, as explained in Sec.~\ref{sec:edm}).
In the visualisation, the brown samples are from the open-set data sets, while all other colours represent the true CIFAR-10 classes.  
This clearly shows that our proposed EDM is quite effective at separating open-set samples from the other clean and closed-set training samples, while DivideMix and ILON largely overfit these samples, as evident from the spread of open-set samples around CIFAR-10 classes.
Interestingly, RoG also shows good separation, but with apparently more complex distributions than EDM.%, which can make training more complex.

% However, the baseline accuracy for DivideMix and EDM when there is no open-set noise i.e. $\omega=1$ is much higher than RoG and ILON, allowing them (DivideMix and EDM) to still be more accurate for large rates of open-set noise.
%where the result for DivideMix and EDM are comparable.  
%Once open-set noise is introduced, the deterioration is more pronounced in the DivideMix result than in our proposed EDM, especially when using CIFAR-100 as open-set noise.
% Finally, another interesting observation is that the EDM accuracy is better than DivideMix for $\rho=0.3$ for all closed-set noise rates $\omega$ and open-set data sets.
% Results with CIFAR-100 and ImageNet32 as open-set data sets are surprisingly similar, even though one would expect that our EDM method would be more challenged by an open-set data set like CIFAR-100, which is more similar to CIFAR-10.

%\gustavo{Hopefully we can have one more result, either the asymmetric noise or t-SNE.}

% \gustavo{compare the 2-component D and 3-component S results. Comment results}

% \ragav{EDIT THIS:} As mentioned in Sec.~\ref{sec:edm}, the main contribution of our proposed EDM in Alg.~\ref{alg:EDM} is extending DivideMix~\cite{DivideMix} to work with the combined open-set and closed-set noise problem, as the one proposed in Sec.~\ref{sec:proposed_benchmark}.
\textbf{Noise classification:} Fig.~\ref{fig:EDMvsDM} shows the distribution of loss function values for the clean, open-set and closed-set samples in cases where the noise rate $\rho=0.6$, with closed-set rates $\omega \in \{0,0.5,1\}$ using samples from both CIFAR-100 and Imagenet32 as open-set noise.
From these graphs, it is clear that the SL loss in Eq.~\eqref{eq:S_loss_sample} is able to successfully distinguish samples from each one of the three sets above, even when only one of the noise types is available, such as the case when $\omega \in \{0,1\}$. 
%\gustavo{this is not really clear from the graph because we don't show the result frm the GMM clustering.}  
This suggests that the exploration of uncertainty in the SL loss to identify samples belonging to open-set noise is effective.
Among the methods tested in this paper, DivideMix~\cite{DivideMix} also tries to separate the training samples into clean and noisy sets using the loss in~\eqref{eq:D_loss_full}. 
However, the resulting distribution seems inadequate to allow for a clear separation between the three sets because the open-set and closed-set noisy labels are basically indistinguishable.  Consequently, DivideMix is able to separate clean samples from noisy samples, but not closed-set noise from open-set noise, thus forcing it to treat both noise types similarly during the training (i.e., both types are treated as closed-set noise). This is not ideal given that the open-set samples will be allocated to one of the incorrect training labels, which can ultimately cause the training to overfit these samples.

\section{Conclusion}

In this paper, we investigate a variant of the noisy label problem that combines open-set~\cite{OpenSetPaper,lee2019robust} and closed-set noisy labels~\cite{DivideMix,lee2019robust}. To test various methods for this new problem, we propose a new benchmark that systematically changes the total noise rate and the proportion of closed-set and open-set noise. The open-set samples were sourced from either a small-scale data set (CIFAR100) or a large-scale data set (ImageNet32) such that the true label of these samples is not contained in the primary data set (CIFAR10).
We argue that such a problem setup is more general and similar to real-life noisy label scenarios. We then propose the EvidentialMix algorithm to successfully address this new noise type with the use of the subjective logic loss~\cite{sl} that produces low loss for clean samples, high loss for closed-set noisy samples, and mid-range loss for open-set samples.
%, and SSL techniques.
The clear division of the training data allows us to (1) identify and thereby remove the open-set samples from training to avoid overfitting them, given that they do not belong to any of the known classes, and (2) learn from the predicted closed-set samples in a semi-supervised fashion as in~\cite{DivideMix}. The evaluation shows that our proposed EDM is more effective to address this new combined open-set and closed-set label noise problem than the current state-of-the-art approaches for closed-set problems~\cite{DivideMix,lee2019robust} and open-set problems~\cite{OpenSetPaper,lee2019robust}.

\textbf{Future work:} The motivation for introducing this problem was to open the dialogue in the research community to investigate the combined open-set and closed-set label noise. Moving forward, %in order to further bridge the gap between real-life noise scenarios and the ones currently being investigated, 
we aim to explore more challenging noise settings such as incorporating asymmetric~\cite{ReedEarly} and semantic noise~\cite{lee2019robust} in the proposed combined label noise problem. 
Since we are the first ones to address this problem in a \textit{controlled} setup, there is no precedent on how these more challenging noise scenarios could be meaningfully incorporated. For instance, even though asymmetric closed-set noise has previously been studied in the literature~\cite{ReedEarly}, it is not obvious what its counterpart,~asymmetric open-set noise entails; for instance, it is not immediately clear  how  to build a noise transition matrix between CIFAR10 and ImageNet classes.
% There are several open-ended questions here - What is the relationship between open-set and closed-set noises in an asymmetric setting? Are the classes that contain closed-set noise more likely to contain open-set noise as well? How does the semantic similarity of open-set images to the class label tie in with the asymmetry? 
%In order to meaningfully emulate real-life noise scenarios, it is imperative to address these questions. As every decision or choice can completely change the nature of the experimental setup, a thorough investigation is required which we consider to be beyond the scope of this paper. 
%We plan to investigate these questions in a future version of our research.
In addition, we see merit in investigating other types of uncertainty to identify open-set noise, such as with Bayesian learning~\cite{gal2016dropout}, and aim to explore such methods. %Finally, we plan to either find existing or propose new curated benchmark data sets that naturally contain a combination of open-set and closed-set noise for future use.

\textbf{Acknowledgements:} IR and GC gratefully acknowledge the support of the Australian Research Council through the Centre of Excellence for Robotic Vision CE140100016 and Future Fellowship (to GC) FT190100525.  GC acknowledges the support by the Alexander von Humboldt-Stiftung for the renewed research stay sponsorship. RS acknowledges the support by the Playford Trust Honours Scholarship.

{\small
\bibliographystyle{ieee_fullname}
\bibliography{egbib}
}

\end{document}